\documentclass[journal]{IEEEtran}
\usepackage{epsfig}
\usepackage{graphicx}
\usepackage{amsmath}
\usepackage{amssymb}
\usepackage{subfigure}
\usepackage{color}
\usepackage{epstopdf}
\usepackage{algorithm}
\usepackage{algorithmic}
\usepackage{multirow}

\usepackage{times}
\usepackage{amsmath}
\usepackage{amssymb}
\usepackage{booktabs}
\usepackage{enumerate}
\usepackage{float}
\usepackage{caption} %
\usepackage{makecell}

\usepackage{hyperref}
\hypersetup{
    colorlinks=true,
    linkcolor=blue,
}

\begin{document}

\title{FineAction: A Fine-Grained Video Dataset for Temporal Action Localization}



\author {
    Yi Liu \textsuperscript{\rm 1} , 
    Limin Wang  \textsuperscript{\rm    2} , 
    Yali Wang  \textsuperscript{\rm 1} , 
    Xiao Ma  \textsuperscript{\rm 1} , 
    and Yu Qiao \textsuperscript{\rm 1,3}\\
    \textsuperscript{\rm 1} Shenzhen Institutes of Advanced Technology, Chinese Academy of Sciences\\
    \textsuperscript{\rm 2} State  Key  Laboratory for Novel Software Technology, Nanjing University\\
    \textsuperscript{\rm 3} Shanghai AI Laboratory\\
    \textsuperscript{}{\{yi.liu1,  yl.wang, xiao.ma, yu.qiao\}@siat.ac.cn, lmwang@nju.edu.cn}
    }



\maketitle

\begin{abstract}
Temporal action localization (TAL) is an important and challenging problem in video understanding. However, most existing TAL benchmarks are built upon the coarse granularity of action classes, which exhibits two major limitations in this task. First, coarse-level actions can make the localization models overfit in high-level context information, and ignore the atomic action details in the video. Second, the coarse action classes often lead to the ambiguous annotations of temporal boundaries, which are inappropriate for temporal action localization. To tackle these problems, we develop a novel large-scale and fine-grained video dataset, coined as {\em FineAction}, for temporal action localization. In total, FineAction contains 103K temporal instances of 106 action categories, annotated in 17K untrimmed videos.
\textcolor{black}{Compared to the existing TAL datasets, our FineAction takes distinct characteristics of fine action classes with rich diversity, dense annotations of multiple instances, and co-occurring actions of different classes, which introduces new opportunities and challenges for temporal action localization.}
To benchmark FineAction, we systematically investigate the performance of several popular temporal localization methods on it, and deeply analyze the influence of fine-grained instances in temporal action localization. 
\textcolor{black}{As a minor contribution, we present a simple baseline approach for handling the fine-grained action detection, which achieves an mAP of 13.17\% on our FineAction.
We believe that FineAction can advance research of temporal action localization and beyond.
The dataset is available at \href{https://deeperaction.github.io/datasets/fineaction}{https://deeperaction.github.io/datasets/fineaction}.
}
\end{abstract}


%
\IEEEpeerreviewmaketitle

\section{Introduction}
Video understanding \cite{kuehne2011hmdb, shao2020intra, Simonyan2014Two, soomro2012ucf101, Du2015Learning, TSN19,yang2020temporal,TDN21} is becoming an important problem in the computer vision community due to its wide applications in video content analysis, intelligent surveillance, and human-computer interaction.
Temporal action localization (TAL) is a fundamental task in this research, which aims to localize the  start and end time of each action instance in the untrimmed videos.
The advance in this area has been mainly driven by deep learning models \cite{Shou2016Temporal,zhao2017temporal,lin2019bmn, xu2020g, lin2020fast} with large-scale benchmarks such as THUMOS14  \cite{idrees2017thumos}, ActivityNet \cite{caba2015activitynet}, and HACS Segment \cite{zhao2019hacs}.

\begin{figure}[t]
\begin{center}
\includegraphics[width=\linewidth]{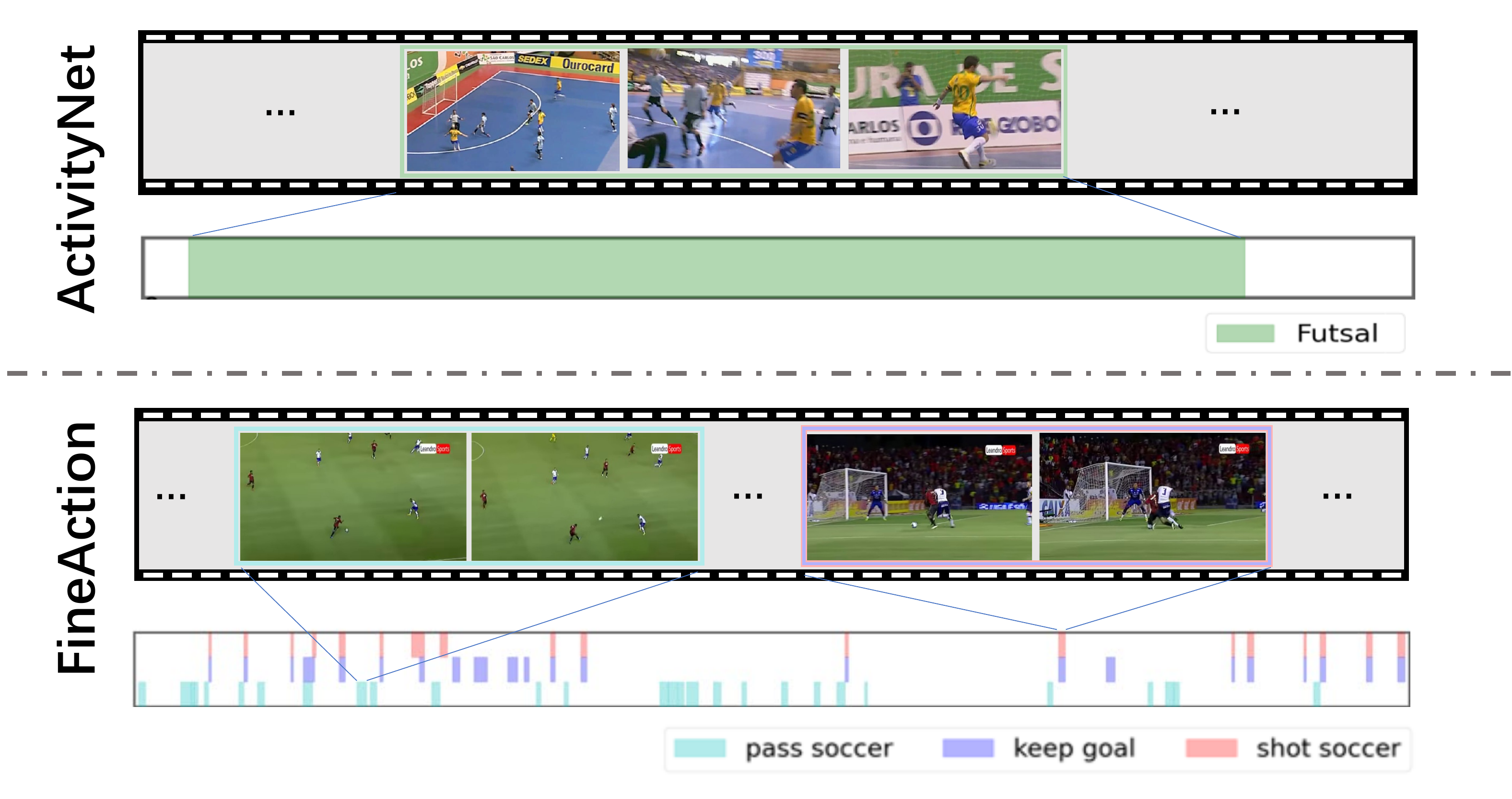}
\end{center}
\caption{Annotation comparison between ActivityNet and FineAction. In the soccer-related videos, the comparison (\textit{Pass soccer}, \textit{Keep goal} and \textit{Shot soccer} v.s. \textit{Futsal}) shows that our FineAction is more fine-grained with rich diversity, rigid and dense annotation of multi-class instances.  }
\label{fig:vision}
\end{figure}

However, the existing benchmarks have several limitations that may hinder the development of TAL methods.
First, some datasets only contain sparse annotations of coarse-level action classes (e.g., ActivityNet and HACS Segment),
where each action instance could occupy a large portion of the whole video.
We argue that this fact fails to reflect the general and realistic requirement of TAL, such as localizing dense action instances in a video, determining more accurate temporal duration for fine-grained actions, handling co-occurring actions of different classes.
As shown in Figure~\ref{fig:vision}, for example, 
ActivityNet contains a single action instance of class \textit{Futsal}, which covers a long duration in the video. We observe that it is inappropriate to localize such high-level event as there is no clear definition on temporal boundary of \textit{Futsal}. Instead we can only accurately localize more fine-grained actions such as \textit{Pass soccer}, \textit{Keep goal} and \textit{Shot soccer} due to its more accurate definition in time.
In addition, detecting these fine-grained actions will pose more challenges to TAL methods with requirement of focusing on action itself rather than background context.
Second, some action datasets  (e.g., THUMOS14) contains small number of action classes with detailed annotation and simply focuses on a specific domain such as sport activities. The small number of training videos would increase the overfitting risk in model development. Meanwhile, it is unclear whether these methods that perform well on sports action temporal localization could be still effective for other daily life actions. Thus, it is expected to increase the scale and diversity of fine-grained TAL datasets to promote the future research.

Based on the discussion above, 
we think it is necessary to establish a new benchmark for TAL by proposing the following criteria.
1) We should focus on more fine-grained actions with clear temporal boundaries, which is more suitable for the task of TAL. Meanwhile, localizing such fine-grained actions can provide more detailed understanding of video content.
2) We should perform relatively dense annotations of fine-grained action classes, possibly assigning multiple labels to overlapped segments, as different fine-grained actions happen simultaneously.
For example, \textit{Keep goal} and \textit{Shot soccer} are coupled in the soccer videos, as shown in Figure \ref{fig:vision}.
These dense annotations will pose new challenges to TAL methods and are more similar to realistic applications such as sports analysis.
3) Instead of working on specific sports actions, this dataset should be large scale and cover various complex human activities in our daily life.

To achieve this goal, we introduce a large-scale and fine-grained video dataset, 
coined as {\em FineAction}, for the task of TAL.
Different from the existing datasets, our FineAction is unique in aspects of fine-grained action classes, multi-label and dense annotations, relatively large-scale capacity, and rich action diversity.
In total, FineAction contains 103K temporal instances of 106 action categories annotated in 17K untrimmed videos.
The whole construction process consists of three critical stages.
First, we define a new class taxonomy with three levels of category granularity.
Our taxonomy excludes coarse-level action definition, and categorizes the highly-related actions into the same subordinate.
As a result,
this taxonomy can effectively organize action categories according to its fine-grained  characteristics in a hierarchical manner.
Second, in order to increase data scale as well as diversity, we progressively collect videos from both existing benchmarks and Internet. 
Third, we develop a customized annotation tool with temporal annotation guidance,
so that annotators leverage image-text illustration to clearly understand fine-grained actions,
and efficiently annotate temporal boundaries of action instances in a video. 
We also conduct annotation quality inspection with multiple rounds of professional checking.
Finally,
we systematically study a number of popular TAL methods on our new benchmark.
Our experiments show that their performance on our FineAction is much worse than other benchmarks, demonstrating that our FineAction brings new challenges for the task of TAL, due to its fine-grained characteristics and detailed annotations.

In summary, FineAction benchmark contributes to the research of TAL in three different ways:
1) The definition of action categories is much more fine-grained with a distinct three-granularity taxonomy. 
2) We provide effective annotation guidance and carefully control the annotation quality, trying to reduce the bias of annotations and the boundary uncertainty of action instances. 
3) 
\textcolor{black}{We propose an initial solution with multi-scale transformer neck to address the fine-grained action localization and  the proposed baseline achieves the mAP of 13.17\% on our FineAction.}
4)
We conduct in-depth studies on FineAction, which reveals the key challenges that arise in the localization of fine-grained instances. 
We expect that our dataset can facilitate the future research of fine-grained TAL.

\begin{figure*}[t]
\begin{center}
\includegraphics[width=0.9\linewidth]{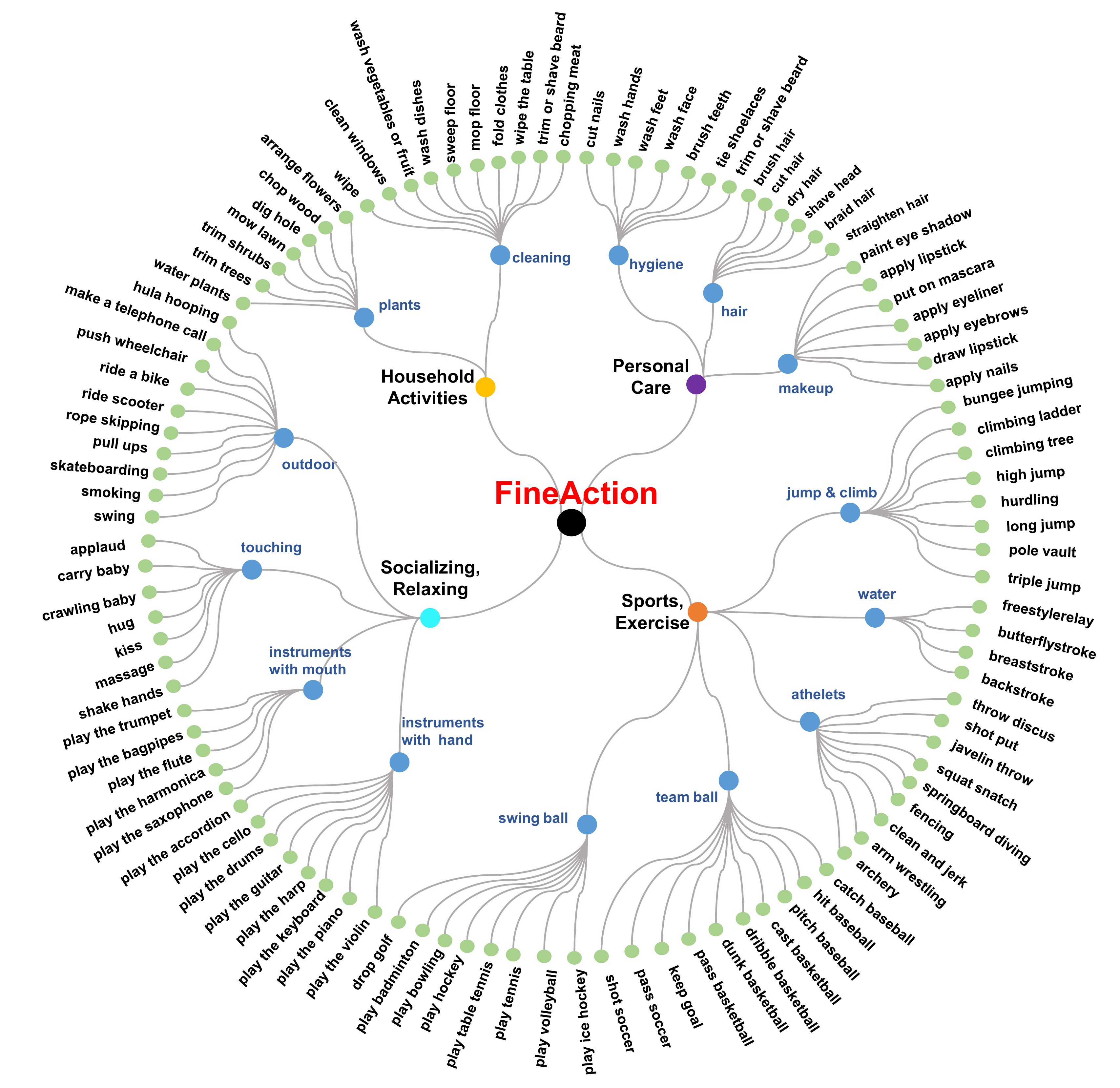}
\end{center}
\caption{Complete organizational taxonomy behind our FineAction, which consists of 4 top-level categories ,14 middle-level categories, and 106 bottom-level categories.}
\label{fig:taxonomy}
\end{figure*}

\begin{figure*}[t]
\begin{center}
\includegraphics[width=0.95\linewidth]{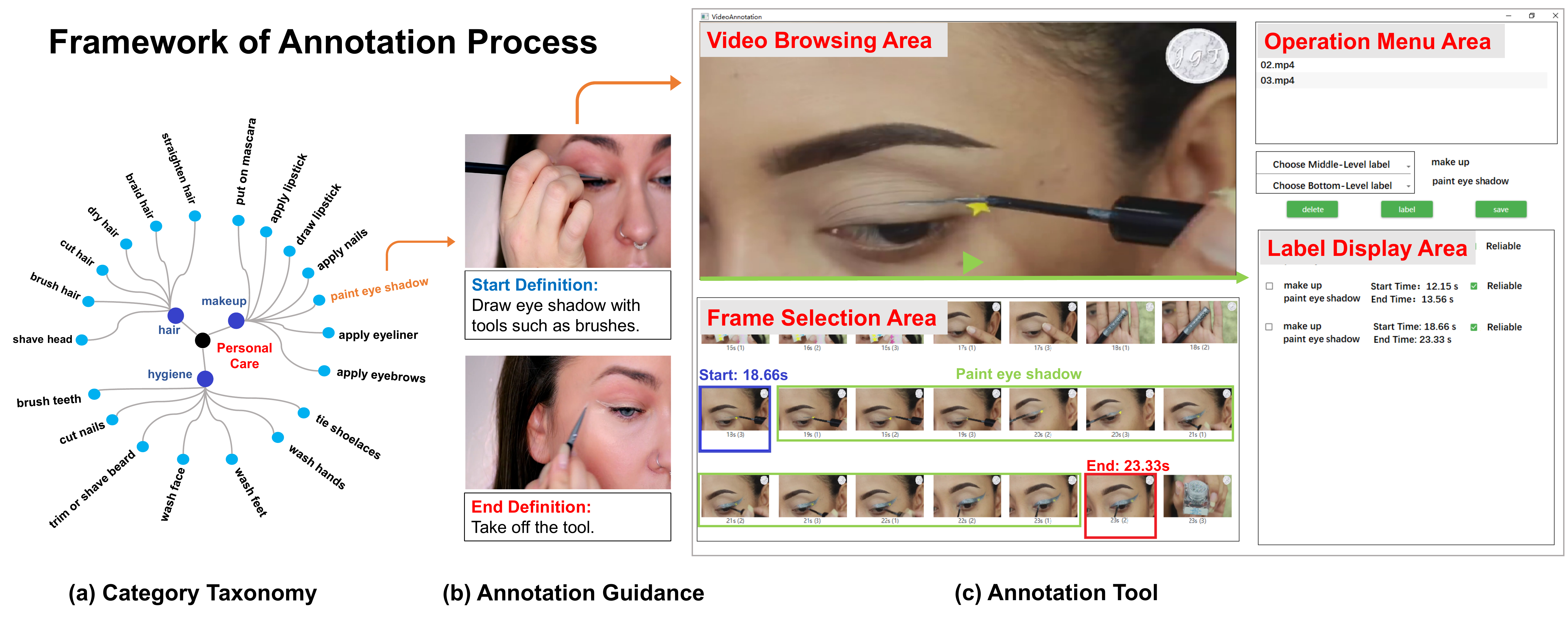}
\end{center}
\caption{The framework of annotation process. (a) The sub-tree of \textit{Personal Care} in Category Taxonomy. (b) An example of \textit{Paint Eye Shadow} for Annotation Guidance. (c) The UI interface of Annotation Tool.}
\label{fig:UI}
\end{figure*}

\section{Related Works}
\subsection{Action Datasets}
Action recognition aims at classifying action in a trimmed video clip.
Due to its potential value in realistic applications, it has been widely explored in the past decades, from handcrafted methods \cite{Laptev2008Learning,wang2013dense} to deep learning models \cite{CarreiraQuo,Du2015Learning,Simonyan2014Two}.
In particular, more challenging datasets were proposed subsequently, including UCF101 \cite{soomro2012ucf101}, Kinetics \cite{CarreiraQuo}, ActivityNet \cite{caba2015activitynet}, Moments in Time \cite{monfort2019moments}, and others \cite{kuehne2011hmdb,idrees2017thumos,yeung2018every,zhao2019hacs,sigurdsson2016hollywood}.
With fast development of these large-scale benchmarks, 
deeply-learned video models exhibit high accuracy by learning spatial and temporal evolution of actions.

Unlike early datasets of action recognition, which mainly focus on action classification, 
more challenging datasets are proposed for other tasks such as temporal action detection and spatio-temporal action detection.
Temporal action detection in untrimmed videos is crucial to inspection of massive videos uploaded on the  Internet  in real world. 
Several datasets with fine granularity of classes have been presented in the narrow domains.
For example,  
THUMOS Challenge 2014 \cite{idrees2017thumos} includes 413 untrimmed videos on 20 sport actions, which was extended into Multi-THUMOS \cite{yeung2018every} with 65 action classes. 
FineGym V1.0 \cite{shao2020finegym} provides temporal annotations with a three-level semantic hierarchy for gymnastic videos.
Other datasets include MPII Cooking \cite{rohrbach2012database,rohrbach2016recognizing} and EPIC-Kitchens \cite{damen2018scaling} mainly focus on scene in the kitchen.
Models trained on such domain-specific datasets may not generalize well to  daily activities.
Conversely, some datasets like ActivityNet-v1.3 \cite{caba2015activitynet} and HACS Segment \cite{zhao2019hacs} were  designed to include more general, every-day activities. 
ActivityNet-v1.3 \cite{caba2015activitynet} includes 20K untrimmed videos with 23K temporal action annotations, and HACS Segment \cite{zhao2019hacs} contains 49k untrimmed videos with 122k temporal action annotations. 
But these datasets are lack of fine-grained annotations for daily activities.

Spatio-temporal action detection datasets, such as UCF Sports \cite{rodriguez2008action}, UCF101-24 \cite{soomro2012ucf101}, J-HMDB \cite{jhuang2013towards}, DALY \cite{weinzaepfel2016towards}, AVA  \cite{gu2018ava} and AVA-Kinetics \cite{li2020ava}, typically evaluate spatio-temporal action detection for short videos with frame-level action  annotations. 
These benchmarks pay more attention to spatial information with frame-level detectors and clip-level detectors, which are limited to fully utilize temporal information. Recently, MultiSports  \cite{MultiSports21} presented a spatio-temporal action detection dataset  on sports actions.
Alternatively,
we mainly focus on temporal localization of detailed human action instances in this work.

\subsection{Temporal Action Localization}
Temporal action localization aims at localizing the start and end points of action clips from the entire untrimmed video with full observation.
Among temporal action detection methods, 
anchor-based action localization methods \cite{lin2017single, long2019gaussian, liu2019multi, gao2017turn, gao2017cascaded,ChaoRethinking,XuR} in a top-down pipeline, try to predict confident of action segments on default anchors and refine the action boundaries.
For example,
\cite{ChaoRethinking,XuR} transforms Faster R-CNN \cite{ren2015faster} for temporal action localization.
On the other hand, actionness-based action localization methods \cite{shou2017cdc, zhao2017temporal, Lin_2018_ECCV, lin2019bmn} in a bottom-up pipeline, which predict the frame-level actionness score to discover actions boundaries.
For example, \cite{zhao2017temporal} leverages actionness as guidance to obtain temporal intervals and then classifies these intervals by learning temporal structure and action completeness.
\cite{Lin_2018_ECCV} generates proposals via learning starting and ending probability using a temporal convolutional network.
Besides, there are methods based on reinforcement learning \cite{yeung2016end}, gated recurrent unit \cite{buch2017sst, zhu2017uncovering} and graph convolutional network \cite{Zeng_2019_ICCV,bai2020boundary}.
Recently \cite{tan2021relaxed} presents a simple and end-to-end learnable framework for direct action proposal generation, by re-purposing a Transformer-alike architecture.

\section{FineAction Dataset}
Our goal is to build a large-scale and high-quality dataset with fine-grained action classes and dense annotations for temporal action localization.
In this section, we present the action taxonomy, the collection and annotation process, the dataset statistics, and the dataset properties.

\subsection{Dataset Preparation}\label{Preparation}
First,
we build up a pool of action categories from the existing benchmarks for video collection and annotation.
Since ActivityNet \cite{caba2015activitynet} and Kinetics \cite{CarreiraQuo} contain a wide range of action categories from sports to daily activities,
we choose the union set of their action categories as our pool to increase class diversity.
Second, to select suitable action classes with clear temporal boundaries, we conduct filtering based on this pool.
Note that, fine-grained classes refer to subordinate-level categories in the superior class, and their inter-class differences are subtle.
Base on this principle,
we carefully define fine-grained action classes with two following cases.
\textbf{(1)} if an action can be decomposed into subordinate-level actions by Wikipedia and/or professional manuals,
it will be chosen as coarse-grained action to be decomposed (e.g., layup is broken down into dribbling, passing ball, dunking).
\textbf{(2)} if several classes are often confused in the research of action recognition,
these classes are also picked as our fine-grained classes due to subtle inter-class differences 
(e.g., play trumpet vs play saxophone, ride a bike vs. ride scooter).

\textbf{Action taxonomy}.
Based on these rules,
we finally generate 106 action classes within a new taxonomy of three-level granularity.
Figure \ref{fig:taxonomy} shows the full organizational taxonomy behind FineAction.
This taxonomy consists of 4 top-level categories (\textit{Household Activities}, \textit{Personal Care}, \textit{Socializing, Relaxing} and \textit{Sports, Exercise}), 14 middle-level categories,
and 106 bottom-level categories.
It is worth noting that,
our built taxonomy is different from the one in ActivityNet \cite{caba2015activitynet}, and HACS Segments \cite{zhao2019hacs},
due to its fine-grained characteristics.
For example,
\textit{Makeup} is actually a middle-level action that consists of 7 bottom-level actions in our taxonomy,
as shown in Figure \ref{fig:UI} (a), while
it is a bottom-level action in ActivityNet. 
In practice, thanks to our more specific action definition, our taxonomy is more reasonable and suitable for temporal action localization in videos.

\textbf{Video collection}.
Based on our action taxonomy,
we progressively collect videos from both existing benchmarks and Internet.
First,
we manually select videos from the related categories of the existing video datasets,
i.e.,
YouTube8M \cite{abu2016youtube}, 
Kinetics400 \cite{CarreiraQuo},
and
FCVID \cite{jiangfcvid}.
Note that,
for the bottom-level action categories that do not appear in these datasets,
we use the corresponding middle-level action categories for selection.
Second,
we annotate these videos (the procedure is described in the next section),
and count the number of instances per category.
If the number of instances is smaller than 200 in one bottom-level category,
we use keyword to crawl YouTube videos to increase instances in this action.
Finally, we remove video duplicates by computing the pair-wise video similarity with deep features.
As for reproducibility,
we will release our dataset with
raw videos \& URLs, 
annotation file,
and 
extracted features. 
But like ImageNet,
our release is based on request,
where researchers must submit the license form and agree to use this dataset for only academic purpose,
before they can access to it.

\begin{table*}[h]
\centering 
\begin{tabular}{c|cccccccc}
\hline
{\textbf{Database}} &{\textbf{Category}}  &\textcolor{black}{\textbf{M-L}}&	{\textbf{Video}} &	{\textbf{Instance}}	&{\textbf{Overlap}}	& {\textbf{Duration}} 	&{\textbf{Action type}} &{\textbf{Main task}}\\
\hline
MPII Cooking \cite{rohrbach2012database}	& 65 &\checkmark	&45 &5,609 & 0.1\%	& 11.1 m	 &kitchens &Action Classification\\
EPIC-Kitchens  \cite{damen2018scaling}	& 4,025 &\checkmark	&700 &89,979 &28.1\%	& 3.1 s	 &kitchens & Action Classification\\
FineGym V1.0 \cite{shao2020finegym}	& 530 &\checkmark & 303	&32,697 &  0.0\% &1.7 s		 &sports & Action Classification\\
\hline
THUMOS14 \cite{idrees2017thumos}	& 20 &$\times$	&413 & 6,316 &17.5\%	& 4.3 s	 & sports & Temporal Action Localization\\
ActivityNet	\cite{caba2015activitynet} & 200 &$\times$ &19,994	& 23,064 &0.0\%	& 49.2 s	 &daily events& Temporal Action Localization\\
HACS Segment \cite{zhao2019hacs}& 200 &$\times$	& 49,485 & 122,304 &0.0\%	& 33.2 s	 & daily events& Temporal Action Localization\\
\textbf{FineAction} & 106 &\checkmark& 16,732 	& 103,324 	& 11.5\%  &  7.1 s 	 & daily events& Temporal Action Localization  \\
\hline
\end{tabular}
\caption{Comparison with Related Benchmarks. 
Compapred to the existing TAL datasets, our FineAction takes distinct characteristics of 
fine action classes with rich diversity, 
dense annotations of multiple instances, 
and
co-occurring actions of different classes, 
which introduces new opportunities and challenges for temporal action localization.(\textcolor{black}{M-L means Multi-Label)} }
\label{tab:compare}
\end{table*}

\begin{table}[t]
\centering
\begin{tabular}{c|cccc|c}
\hline
\textbf{Database} & {\textbf{0-2 s}} & {\textbf{2-6 s}} &  {\textbf{6-15 s}} &  {\textbf{\textgreater 15 s}} &  {\textbf{Ins / Vid}}\\
\hline
THUMOS14 \cite{idrees2017thumos} & 2,029 & 2,753 & 1,437 & 99 & 15.29 \\
ActivityNet \cite{caba2015activitynet} & 900 & 3,253 & 4,426 & 14,485&1.15\\
HACS Segment  \cite{zhao2019hacs} &   8,874 & 29,644  &   31,982 & 51,804 &2.47 \\
\textbf{FineAction} & \textbf{66,890} & 15,253 & 10,523 & 10,586 & \textbf{6.17}\\
\hline
\end{tabular}
\caption{Comparison of instance duration. FineAction contains much more short-duration action instances than the one in ActivityNet and HACS Segment for 0-2s instances.}
\label{tab:time}
\end{table}


\subsection{Dataset Annotation}

After data collection,
we perform manual annotation of action instances for each video.
Specifically,
given a video,
the trained annotator is asked to find the start and the end of each action instance as well as associating action tag with this instance.
To boost annotation efficiency and consistency,
we design an annotation tool as shown in  Figure \ref{fig:UI} (c).
First,
the annotator loads a video into \textit{video browsing area},
for quickly previewing the entire video.
Second,
the annotator carefully checks the video frames in \textit{frame selection area}, 
and determines the start and the end of each action instance.
Third,
the annotator operates various buttons (e.g., adding, deleting, and modifying labels) of \textit{operation menu area} for detailed annotation.
Finally,
the annotator makes a double check for the annotation results on \textit{label display area}.

We design a rigorous guidance and checking procedure to ensure the annotation quality.
First,
we develop a guide to give the specific definition of each action class, which can ensure the annotation consistency of temporal boundaries among different annotators.
Specifically, we provide the start and end description of each action category with image-text definition.
For sport activities,
we use their definition and explanation from Wikipedia and/or professional sport manuals.
For other daily activities,
a team of  professional researchers in TAL are asked to summarize the clear boundary definitions to avoid annotation bias. 
Figure \ref{fig:UI} (b) shows an example of\textit{ Paint Eye Shadow} in our annotation guide.
Second,
we arrange another annotator team to check all the annotated videos in multiple rounds to increase annotation quality.
\textcolor{black}{
Specifically,
this team will calculate the annotation statistics of each category from different annotators, 
including 
the average frequency of the action, 
the average duration of the action, 
etc. 
If different annotators have large statistical variations for the same category, 
it means that this category tends to be controversial in the annotation process,
due to cognitive bias between different annotators. 
In this case,
our evaluation team will double-check all the boundary annotations of this category,
by discussing these videos with professional TAL researchers.
}
Finally,
the professional researchers are asked to make the sampling inspection to further alleviate inaccurate and/or inappropriate annotations.

\subsection{Dataset Statistics}

We first compare our FineAction with the large-scale datasets in TAL,
i.e.,
ActivityNet \cite{caba2015activitynet} and HACS Segment \cite{zhao2019hacs}.
As shown in Table  \ref{tab:compare},
FineAction has the comparable data size but exhibits more fine-grained properties than ActivityNet and HACS Segment.
\textcolor{black}{For example, our FineAction has multi-label and dense annotations rather than one instance per video and the fine-grained actions often happen in a short period.}

Hence,
the average duration of  temporal instance in our FineAction (7.1s) is much shorter than the one in ActivityNet (49.2s) and HACS Segment (33.2s).
Moreover,
we further make comparison in terms of instance-duration  distribution in Table \ref{tab:time}.
The distribution of our FineAction is different from that of ActivityNet and HACS Segments,
i.e.,
FineAction contains much more short-duration  action instances (66,890) than the one in ActivityNet (900) and HACS Segment (8,874) for 0-2s instances.
On one hand, it requires much more effort to label such elaborate annotations.
On the other hand,
it brings much larger challenge for temporal action localization.
Note that,
the average number of action instances per video is  
1.15/2.47/6.17 for ActivityNet/HACS Segment/ FineAction.
Hence, our FineAction is more densely annotated among these large-scale datasets.
Finally,
several fine-grained actions can happen simultaneously.
Hence,
11.5\% of temporal segments have multiple action labels with overlaps in our FineAction,
while all the temporal segments are tagged with a single label in ActivityNet and HACS Segment.
Hence, these multi-label annotations would greatly increase the challenge of our FineAction.

Then, we compare our FineAction with other fine-grained temporal datasets.
As shown in Table \ref{tab:compare},
most of these existing benchmarks focus on narrow domains such as sports (THUMOS14 \cite{idrees2017thumos} and FineGym V1.0 \cite{shao2020finegym}) or kitchen activities (MPII Cooking  \cite{rohrbach2012database} and EPIC-Kitchens \cite{damen2018scaling}),
while our FineAction contains multiple types of activities with rich diversity.
In addition, the recent large-scale fine-grained datasets mainly work on other video tasks,
e.g., action recognition (FineGym V1.0) and anticipation (EPIC-Kitchens).
THUMOS14 is the most similar dataset to our FineAction on TAL.
However, this dataset is quiet small with limited number of videos and action instances.
These comparison demonstrates that our FineAction is unique with distinct properties of fine-grained action classes, multi-label and dense annotations, relatively large-scale capacity, and rich action diversity.

\begin{figure*}[t]
\begin{center}
\includegraphics[width=0.95\linewidth]{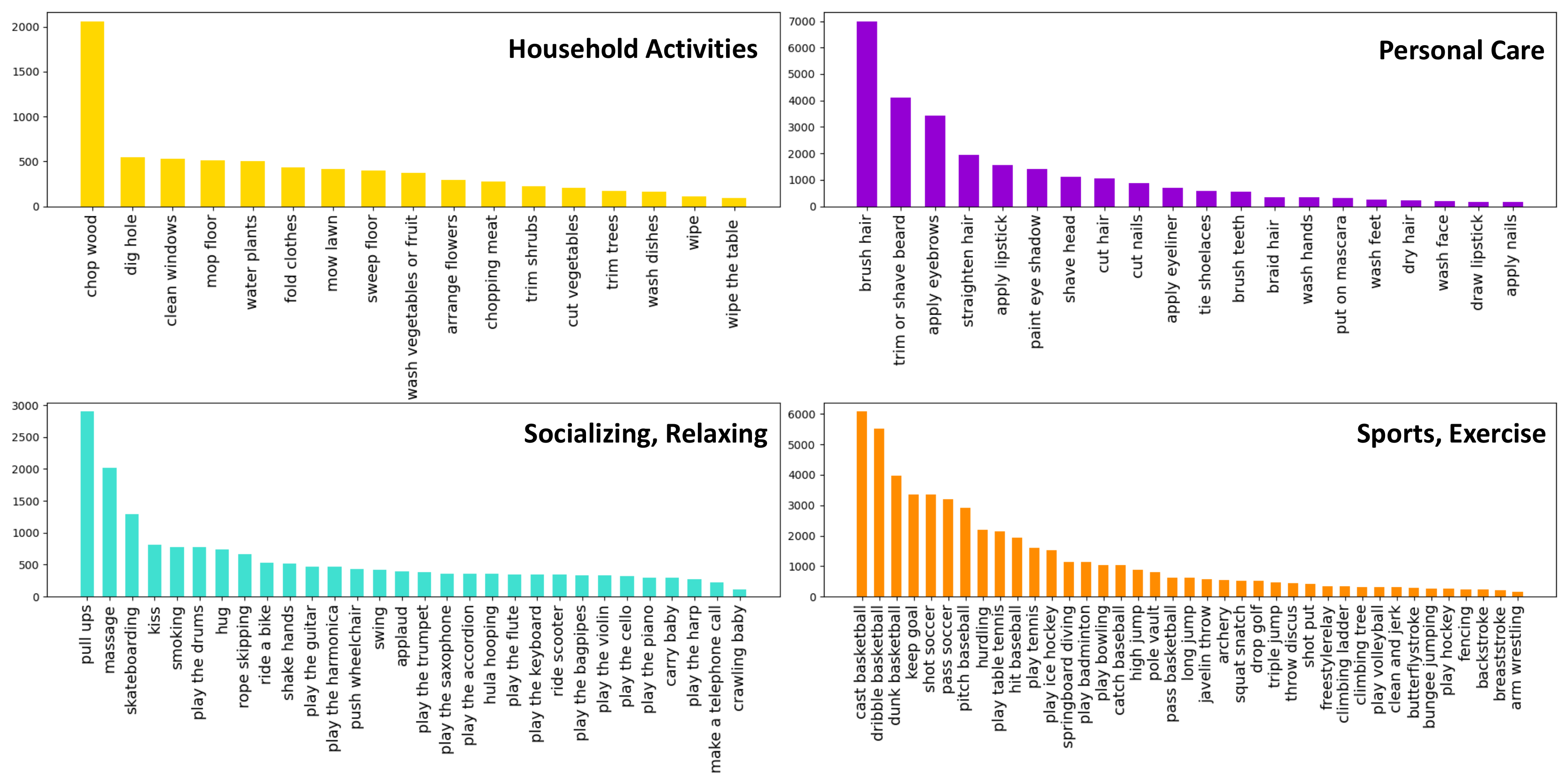}
\end{center}
\caption{Number of instances per category. We plot the instance distribution of all the bottom-level categories in each top-level category. All the plots exhibit the natural long-tailed distribution.}
\label{fig:category_num}
\end{figure*}

Finally,
we show the number of instances for each action category  in Figure \ref{fig:category_num}.
For each top-level category,
we visualize the instance distribution over the corresponding bottom-level categories.
The number of instances exhibits the natural long-tailed distribution,
which also arises new challenges and opportunities for temporal action detection.

\subsection{Dataset Properties}

\textbf{Dataset challenges}. 
Compared with the existing benchmarks,
our FineAction dataset shares the following distinct challenges.
1) {\em Fine-grained action categories with a long-tailed distribution}.
It is difficult to recognize these detailed actions,
if the models mainly focus on distinguishing  background and action categories without learning subtle motion patterns. 
2) {\em  Densely annotated instances with short temporal duration}.
It leads to the difficulties in precisely localizing temporal boundaries,
since various short actions are densely distributed in the entire video.
3) {\em Temporal segments with multiple action labels in overlaps}.
It is challenging to recognize different actions in a single segment,
if the models are designed without concurrent action analysis and/or action relation learning.
4) {\em Diverse action classes with a wide range of semantics}.
Without specific prior knowledge, it is hard to design TAL models to perform well on various fine-grained human actions.

\textbf{High quality}.
FineAction is progressively annotated in a careful and efficient procedure.
Our annotation teams are rigidly supervised by professional researchers.
With our customized annotation tool and guidance,
action instances are correctly labeled with precise temporal boundaries.
Besides,
our researchers make multiple calibrations to remove annotation bias as much as possible.

\textbf{Richness and diversity}. 
From the perspective of action definition,
FineAction establishes a new taxonomy which contains fine-grained actions from sports to daily activities.
Hence, it is a preferable benchmark to understand detailed human actions in the realistic scenarios.
From the perspective of data collection,
FineAction is collected from various data resources 
via rigid quality control.
Such richness brings new opportunity to develop powerful deep learning models for TAL.

\section{Experiment}
Our goal is to show challenges and values of FineAction in TAL research. 
Hence, 
we mainly focus on systematical investigation of widely-used SOTA approaches in this paper. 
Via extensive ablation studies and rigorous error analysis, 
we aim to sufficiently explore core contribution of our FineAction, 
and provide insightful findings that are beneficial for future research of TAL.
\textcolor{black}{
Due to the quality of video features can seriously affect the performance of TAL, as our proposed TAL network takes video features as input. We firstly carried out the experiment of untrimmed video classification task by using different vision backbone.}

\subsection{Dataset Split}\label{Evaluation}

\textbf{FineAction benchmark}. 
To build a solid TAL benchmark, 
we manually split the videos into the training set (50$\%$), validation set (25$\%$), and testing set (25$\%$). 
In total, FineAction contains 57,752 training instances from 8,440 videos, 
24,236 validation instances from 4,174 videos and  21,336 testing instances from 4,118 videos.
Unless otherwise mentioned, 
we report the results by training on the training set and testing on the validation set.
We withhold the annotations of testing set in the public release and encourage to submit the testing results to our server for evaluation.

\begin{table*}[t]
\centering
\begin{tabular}{c|c|cccc|cccc}
\hline
\multirow{2}{*}{\textbf{Method}} & \multirow{2}{*}{\textbf{Modality}}  &\multicolumn{4}{c}{\textbf{ Action Proposal Generation}} &\multicolumn{4}{|c}{\textbf{  Temporal Action Localization}}\\
& &  \textbf{AR@5} & \textbf{AR@10} & \textbf{AR@100}& \textbf{AUC}  & \textbf{mAP@0.50}& \textbf{mAP@0.75}& \textbf{mAP@0.95}& \textbf{Avg.mAP} \\
\hline
\multirow{3}{*}{BMN \cite{lin2019bmn}} &  RGB    &8.62 &11.20 & 22.74  & 17.49 &12.56& 7.49 &2.62 & 7.86\\
&  Flow  &9.85 &12.72 & 24.18  & 18.94 & 14.49 & 8.92 & 3.19 & 9.23\\
&  RGB+Flow   &9.99 &12.84 & 24.34  & 19.19 & 14.44 &  8.92 & 3.12 &9.25\\
\hline
\multirow{3}{*}{DBG \cite{lin2020fast} } &  RGB   &6.82 & 9.01& 21.26 & 15.48 &8.57 & 5.01 & 1.93 & 5.31\\
&  Flow   &8.27 & 10.90 &23.37  & 17.70 & 11.03 &  6.95 & 2.70 & 7.20\\
&  RGB+Flow  & 7.82  & 10.45 &23.07 & 17.24 & 10.65 & 6.43 &  2.50 & 6.75\\
\hline
\multirow{3}{*}{G-TAD \cite{xu2020g} } &  RGB    &7.96 &10.45 &20.86 &16.06 & 10.88 & 6.52 & 2.19 & 6.87\\
&  Flow  &8.87 &11.60 & 22.01 & 17.09 & 12.58 & 8.18 &  2.56 & 8.26\\
&  RGB+Flow  &9.02  &11.83 & 23.17  & 17.65 &  13.74 & 8.83 & 3.06 & 9.06\\
\hline
\textcolor{black}{Ours}&  RGB+Flow &11.47 &15.10 & 29.61 & 23.07 &  22.01 & 12.09 & 3.88 & 13.17\\
\hline
\end{tabular}
\caption{Comparison of the state-of-the-art methods on the validation set of FineAction. \textbf{Left}: evaluation on action proposal generation, in terms of AR@AN. \textbf{Right}: evaluation on action detection, in terms of mAP at IoU thresholds from 0.5 to 0.95.}
\label{tab:SOAT}
\end{table*}

\subsection{\textcolor{black}{Untrimmed Video Classification}}

\textcolor{black}{
\textbf{Evaluation metrics.} As for the untrimmed video classification task with multi-label, we choose Mean Average Precision (mAP)  as the main evaluation metric.
}

\textcolor{black}{
\textbf{Results. }
Based on our FineAction benchmark,
it is natural to introduce a classification task on untrimmed videos.
In this work,
we apply the widely-used video models for task evaluation,
including TSN \cite{wang2016temporal},
I3D \cite{CarreiraQuo}, 
SlowFast \cite{feichtenhofer2019slowfast},
and Video Swin Transformer \cite{liu2021Swin}.
The results of each model can be seen in Table \ref{tab:untrimmed_video}. 
First,
the performance is getting better as the model becomes stronger,
e. g.,
Video Swin Transformer achieves the best classification performance,
with the powerful transformer architecture.
Second,
the overall performance is still weak for this untrimmed video classification task,
due to multi-label and fine-grained action clips in FineAction.
}

\begin{table}[!htp]
\centering
\begin{tabular}{c|cc|c}
\hline
\textbf{Model}  & \textbf{Backbone} & \textbf{clip\_len}  & \textbf{mAP(\%)}\\
\hline
TSN \cite{wang2016temporal}       &   ResNet-50   & 1&      56.28\\
I3D \cite{CarreiraQuo}       &   ResNet-50   &   16&   65.23\\
SlowFast \cite{feichtenhofer2019slowfast}  &  ResNet-50     & 32&66.92 \\
Video-SwinB  \cite{liu2021Swin}  &  Swin-Base    &  32&69.24    \\
\hline
\end{tabular}
\caption{\textcolor{black}{Comparison between different backbones for untrimmed video classification Task on the FineAction.} }
\label{tab:untrimmed_video}
\end{table}

\textcolor{black}{
\textbf{Implementation Details}.
All these models are implemented by the MMAction2  \cite{2020mmaction2} framework. 
For training,
we initialize the network weights from the K400 \cite{CarreiraQuo} pretain models and employ an AdamW \cite{kingma2014adam} optimizer for 30 epochs using a cosine decay learning rate scheduler and 2.5 epochs of linear warm-up. A batch size of 64 is used.
For inference, we follow \cite{arnab2021vivit} by using 10x3 views, where a video is uniformly sampled in the temporal dimension as 10 clips, and for each clip, the shorter spatial side is scaled to 224 pixels and we take 3 crops of size 224x224 that cover the longer spatial axis.
}

\subsection{Temporal Action Localization}
\textbf{Evaluation metrics.} 
In line with the previous temporal action localization tasks, we use Average Recall (AR) to evaluate the proposals on action categories, which is calculated under different tIoU thresholds  [0.50:0.05:0.95].
We measure the relation between AR and the Average Number (AN) of proposals, denoted
as AR@AN. 
Furthermore, we also calculate the area (AUC) under the AR vs. AN curve as another evaluation metric, similar to ActivityNet dataset, where AN ranges from 0 to 100.
For the temporal action localization task, mean Average Precision (mAP) is a conventional evaluation metric, where Average Precision (AP) is calculated for each action category. 
For our FineAction, mAP with tIoU thresholds [0.50:0.05:0.95] is used to compute the average mAP.

\textcolor{black}{
\textbf{Baseline Approach.}
To learn fine-grained action proposals,
we introduce a simple but effective multi-scale transformer for FineAction localization.
First, given an untrimmed video $V$, we can extract a visual feature sequence $F\in \mathbb{R}^{ T\times{D} }$.
Before the multi-scale transformer layers, we use two 1D convolutional network with ReLU to extract the embedded feature $F^{1}$ as the input of transformer layer.
Second,
we introduce five transformer layers as the neck architecture, apply LayerNorm (LN) before every block, and add the residual connection after every block. 
Specifically, we use 2x down-sampling by 1D convolution between each transformer layer to capture the multi-scale features from different temporal scales.
This is given by:
\begin{equation}
\begin{aligned}
&  {F}^{\ell+1}= \Downarrow ( {Transformer}_{\ell} (F^{\ell}) ),\quad \ell=1 \ldots L-1\\
\end{aligned}
\end{equation}
where ${F}^{\ell} \in \mathbb{R}^{T^{\ell}\times{D}}$ and $\Downarrow$ is the 2x down-sampling with 1D convolution.
Finally,
we follow BMN \cite{lin2019bmn} to design the head architecture with Temporal Evaluation Module and Proposal Evaluation Module.
In the inference phase, 
we fuse the boundary probability and the confidence map from multi-scale predictions,
for generating the final confidence scores and proposals.
The whole architecture is shown in Table \ref{tab:architecture}.
}

\begin{table}[t]
\centering
\begin{tabular}{c|cccc}
\hline
\textbf{Module}  & \textbf{Layer}  & \textbf{Input} & \textbf{Output}\\
\hline

\multirow{7}{*}{\textbf{Neck}}        &   1D Convolution1   &       T$\times$D & T$\times$512\\ 
&    1D Convolution2   &       T$\times$512 & T$\times$512\\ 
&   Transformer Unit1   &       T$\times$512 & T$\times$512\\
&   Transformer Unit2   &       T/2$\times$512 & T/2$\times$512\\
&   Transformer Unit3   &       T/4$\times$512 & T/4$\times$512\\
&   Transformer Unit4   &       T/8$\times$512 & T/8$\times$512\\
&   Transformer Unit5   &       T/16$\times$512 & T/16$\times$512\\
\hline
\multirow{2}{*}{\textbf{\makecell[c]{Head}}}  &  TEM \cite{lin2019bmn}& T\_{List}$\times$512 & T\_{List}$\times$2\\ 
&   PEM \cite{lin2019bmn}  & T\_{List}$\times$512 & T\_{List}$\times$2$\times$D\\  
\hline
\end{tabular}
\caption{\textcolor{black}{The architecture of our baseline approach with multi-scale transformers. T\_{List} means $[T,T/2,...,T/16]$ }}
\label{tab:architecture}
\end{table}

\begin{figure}[t]
\begin{center}
\includegraphics[width=0.9\linewidth]{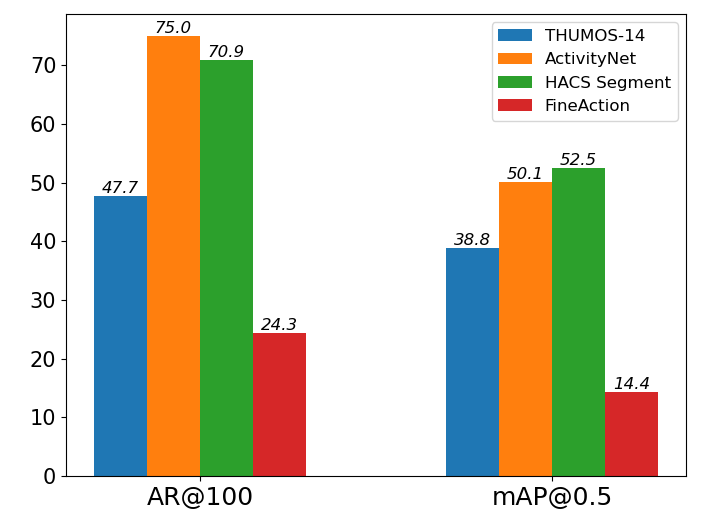}
\end{center}
\caption{ Performance evaluation of BMN method on different benchmarks for temporal action proposal generation and temporal action detection.}
\label{fig:SOAT comparation}
\end{figure}

\textbf{Results. }
We report the performance of the state-of-the-art methods whose codes are publicly available, including BMN \cite{lin2019bmn}, DBG \cite{lin2020fast}, G-TAD \cite{xu2020g} and our baseline approach, on two main tasks for temporal action localization. 
To ensure a fair comparison, we keep the video features on the same scale for all the methods.
Table \ref{tab:SOAT} left shows the results on \textit{Action Proposal Generation}, in terms of AR@AN with SNMS at IoU thresholds [0.50:0.05:0.95], where SNMS stands for Soft-NMS.
Table \ref{tab:SOAT} right shows the results of \textit{Action Detection}, in terms of mAP at IoU thresholds [0.50:0.05:0.95].
To assign the global classification results to the proposals, we adopt top-1 video-level classification results generated by TSN \cite{TSN19}, and we use confidence scores of BMN proposals for detection results retrieving.
It clearly shows that, the performances of these existing methods on our FineAction for the two tasks are comparable to each other, and the modality between RGB and Flow has a certain complementarity.
\textcolor{black}{
Moreover, our proposed baseline achieves the mAP of \textbf{13.17\%} and this further demonstrates the advantage of the multi-scale transformer neck on distinguishing fine-grained action.}
To further compare our FineAction with other benchmarks, we present the performance evaluation of BMN method on different benchmarks in Figure \ref{fig:SOAT comparation}.
The performance on our dataset is far lower than other benchmarks for both action proposal generation and action detection, demonstrating the challenges of our FineAction dataset.

\textbf{Implementation Details. }
For feature encoding, we utilize two-stream architecture \cite{Simonyan2014Two} to extract features for video frames. 
In our experiment, two separate I3D \cite{CarreiraQuo} models are pre-trained from consecutive RGB frames and optical Flow frames on Kinetics \cite{CarreiraQuo}, respectively. 
I3D model takes non-overlapping snippets of 16 stacked RGB or optical Flow frames as input and extracts 2048-dimensional features for each stream. 
For the fusion of modality, we concatenate the features of RGB and Flow.
By default, we rescale the feature sequence of input videos to length $L$ = 100 by linear interpolation following \cite{Lin_2018_ECCV}. We also investigate the effect of rescaled length in our ablation study.
For hyper parameter settings,
we have investigated different parameter settings of TAL methods.
For three methods, L ( window length) is 100, D (maximum duration length) is 50 and SNMS threshold is 0.65.
For DBG,  $\epsilon$ of Gaussian function is 0.75.
For G-TAD, $\tau_1$ and $\tau_2$ are 16 and 16.

\subsection{Ablation Studies}
\textcolor{black}{
\textbf{Study on cross-dataset evaluation}. 
We conduct cross dataset evaluation for temporal action localization. 
Specifically,
we choose BMN \cite{lin2019bmn} as the basic architecture.
Subsequently,
we use ActivityNet to pretrain BMN,
and use FineAction to fine-tune it.
Alternatively,
we also use FineAction to pretrain BMN,
and use ActivityNet to fine-tune it.
The results are shown in Table \ref{tab:cross-dataset} .
As expected,
the mAP performance drops for both cases,
compared with learning from scratch.
The main reason is that,
our FineAction consists of fine-grained actions,
while 
ActivityNet contains coarse-grained actions.
They exhibit different data characteristics with domain shift.
This brings opportunities and challenges for our FineAction benchmark.
}

\begin{table}[t]
\centering
\begin{tabular}{c|cccc}
\hline
\textbf{Dataset}  &\textbf{Parameters Initial} & \textbf{AUC(\%)} \\
\hline
ActivityNet  \cite{caba2015activitynet}       &   Scrath    &  67.29  \\
FineAction         &   Scrath     &  19.19 \\
\hline
ActivityNet   \cite{caba2015activitynet}    &        FineAction  &   63.17 ($\downarrow$ 4.12)\\
FineAction     &       ActivityNet \cite{caba2015activitynet}  &   18.42 ($\downarrow$ 0.77)  \\
\hline
\end{tabular}
\vspace{3mm}
\caption{\textcolor{black}{Study on cross-dataset evaluation on  FineAction with BMN method. }}
\vspace{-4mm}
\label{tab:cross-dataset}
\end{table}

\begin{figure}[t]
\begin{center}
\includegraphics[width=0.95\linewidth]{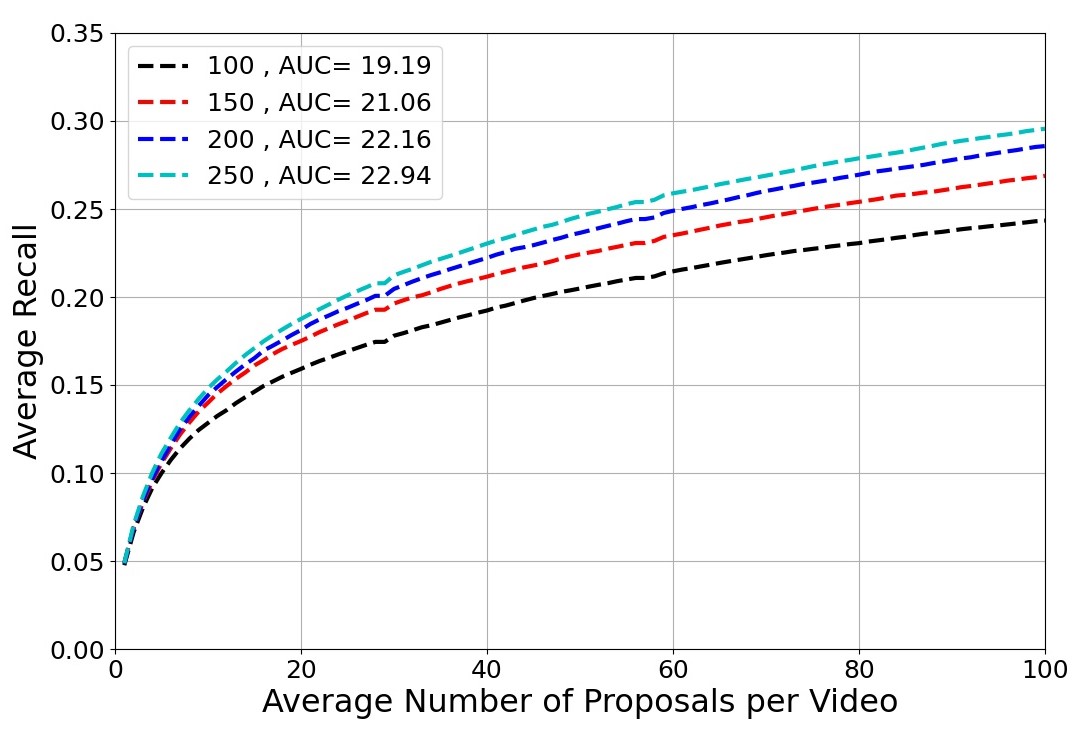}
\end{center}
\caption{ Influence of feature sequence length for Localization performance AR@100  with tIoU thresholds [0.50:0.05:0.95].}
\vspace{5mm}
\label{fig:length}
\end{figure}

\textbf{Study on sequence length}.
We investigate the length of input feature sequence on the FineAction dataset.  
We use BMN \cite{lin2019bmn} as baseline for evaluation and the result is reported in Figure \ref{fig:length}.
As expected, increasing the sequence length contributes to a better localization performance when $L$ is from 100 to 200, and the performance tends to saturate when $L$ is from 200 to 250.
We analyze that BMN can leverage longer sequence to aggregate more detailed context for localization.
However, too long sequence may limit the batch size and has a side effect on network training.
Hence, appropriately increasing sequence length can improve fine-grained temporal action localization.

\begin{figure}[t]
\begin{center}
\includegraphics[width=\linewidth]{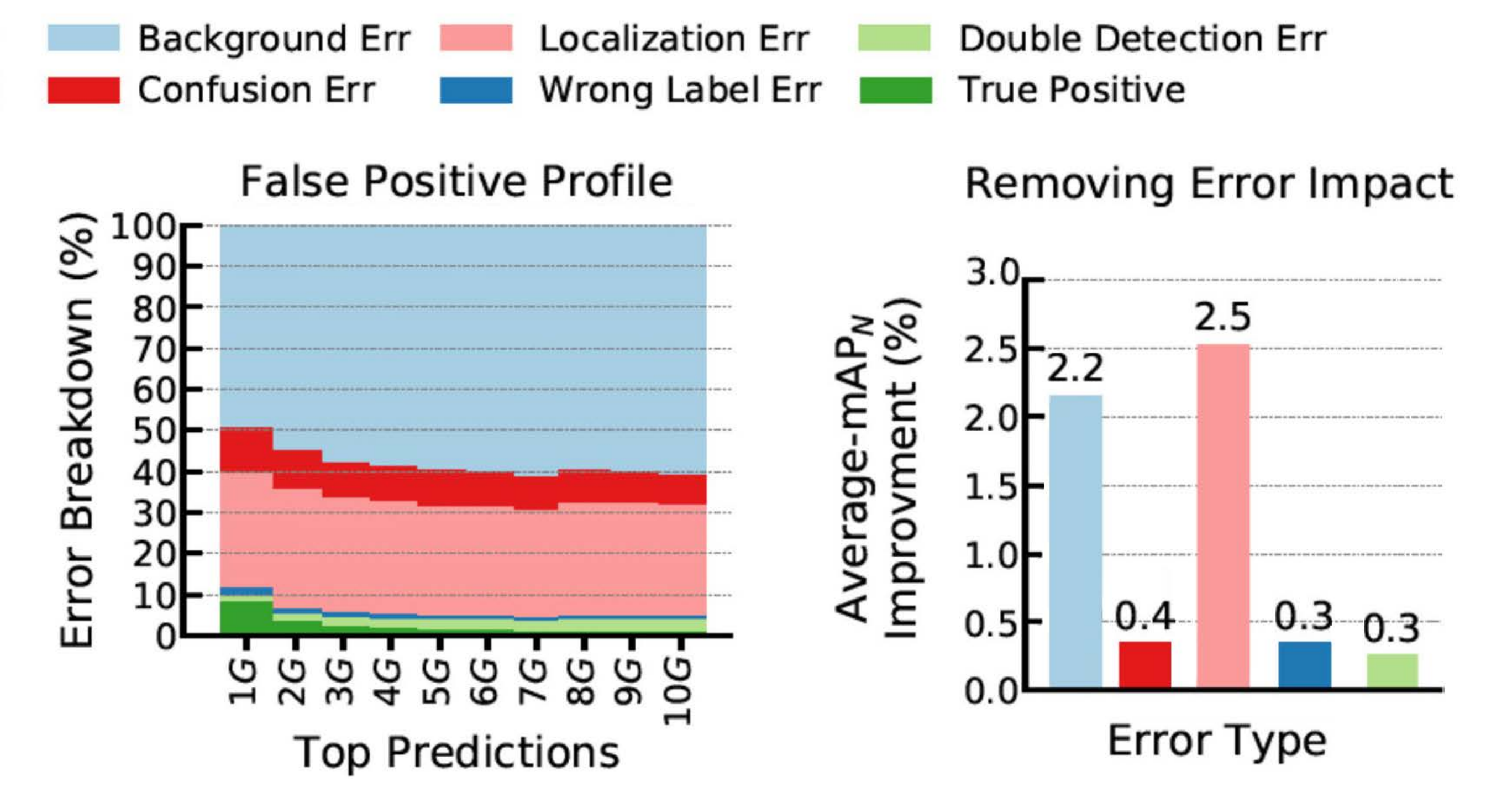}
\end{center}
\vspace{2mm}
\captionof{figure}{ Error analysis on FineAction. \textbf{Left}: the error distribution over the number of predictions per video. G means the number of Ground-Truth instances. \textbf{Right}: the impact of error types, measured by the improvement gained from resolving a particular type of error.}
\vspace{-2mm}
\label{fig:error}
\end{figure}

\begin{figure*}[t]
\begin{center}
\includegraphics[width=.9\linewidth]{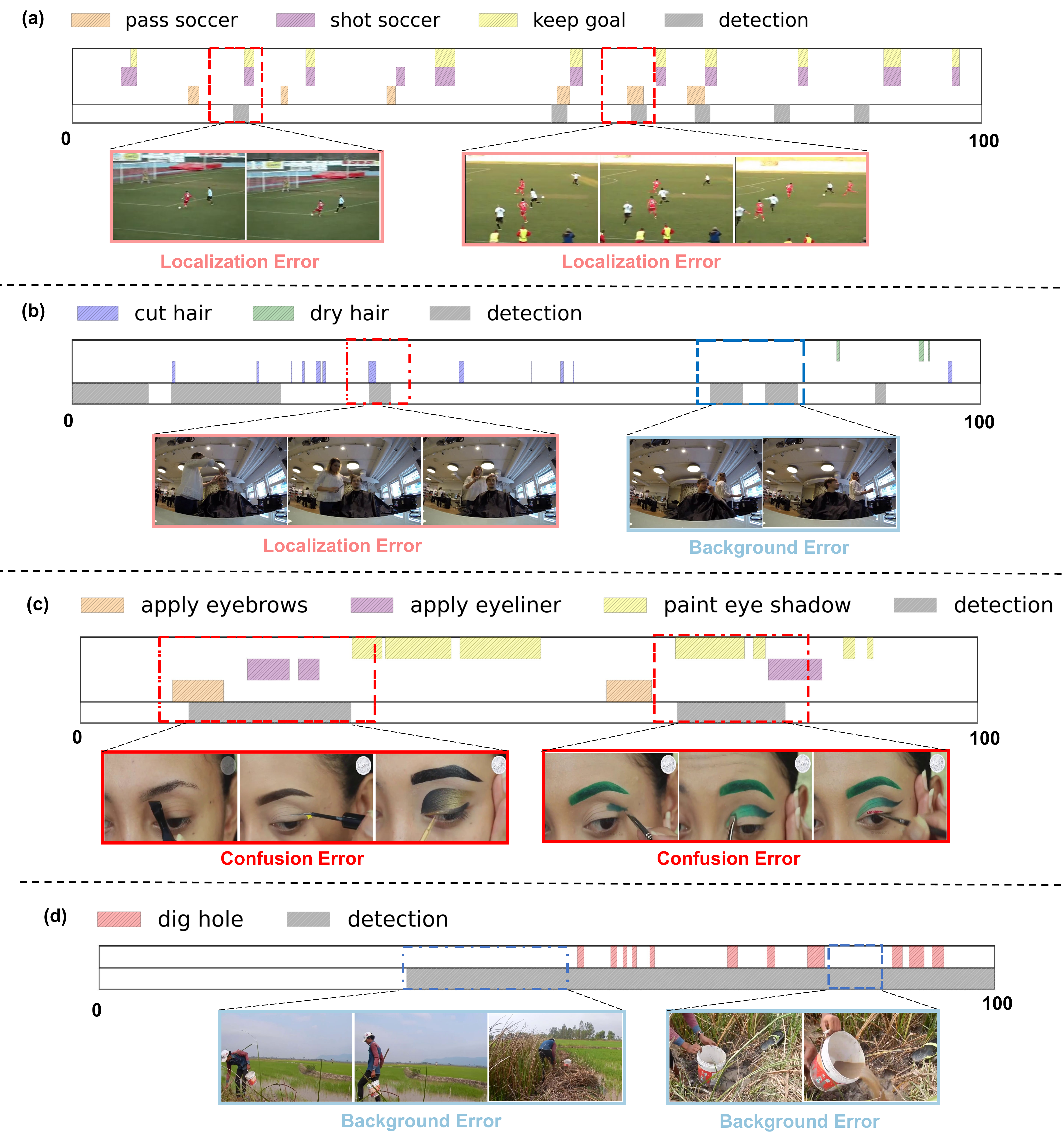}
\end{center}
\caption{Visualizations of typical errors on our FineAction. Background Error and Localization Error are the two main sources of false positives due to the fine-grained and multi-label annotations of our FineAction.   }
\vspace{-4mm}
\label{fig:vis}
\end{figure*}


\textbf{Error analysis.}
We analyze typical errors in FineAction by using BMN method with the protocol in \cite{alwassel2018diagnosing}.
As shown in Figure  \ref{fig:error},
Background Error and Localization Error are the two main sources of false positives.
We can conclude that the BMN method often generates invalid proposals and incorrect boundaries in our FineAction,
perhaps due to the fine-grained and multi-labeled action instances.
To further demonstrate it, we visualize the typical failure cases in Figure \ref{fig:vis}.
Note that, Background Error and Localization Error are the two main sources of false positives.
We can conclude that many detected proposals are invalid and fail to distinguish action from background, which confirms the challenges of our FineAction dataset.



\begin{figure}[t]
\begin{center}
\includegraphics[width=.9\linewidth]{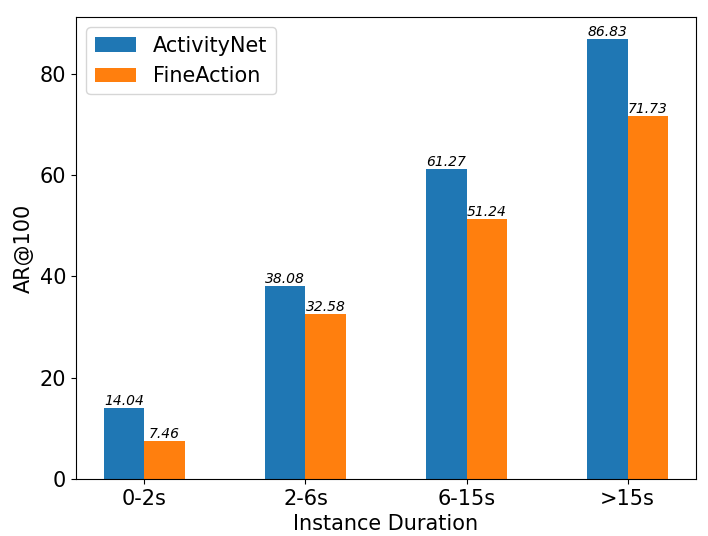}
\end{center}
\captionof{figure}{ Localization performance AR@100 on different duration of instances. Our FineAction is more challenging than ActivityNet in different durations. }
\label{fig:error_time}
\end{figure}
\begin{figure}[t]
\begin{center}
\includegraphics[width=0.95\linewidth]{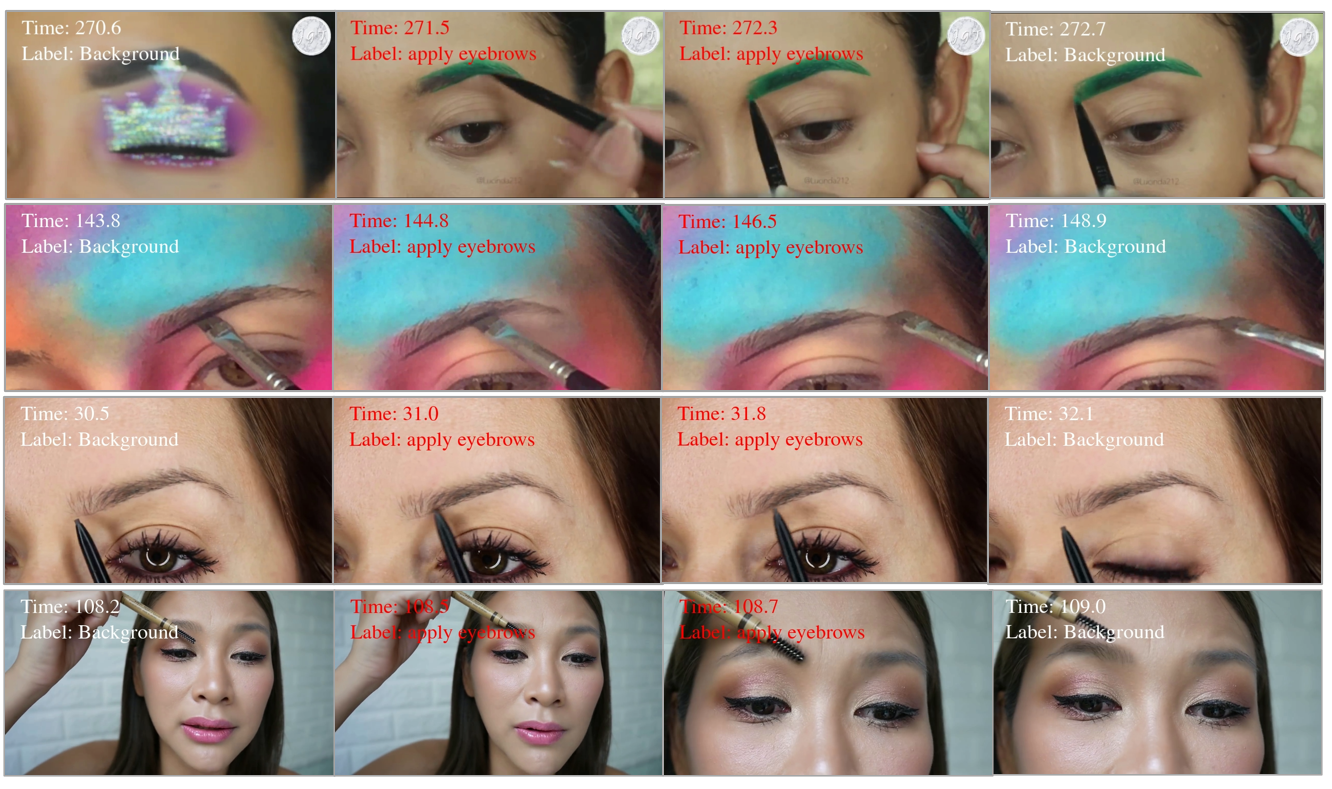}
\end{center}
\caption{ Visualization the video clips for the challenging category of \textit{apply eyebrows}.}
\label{fig: apply_eyebrow}
\end{figure}

\textbf{Which categories are more challenging?}
After carefully analyzing the typical errors, we delve into which categories are more challenging in our FineAction.
First, we divide the instance into four levels according to its duration, 
followed by \textit{0-2s, 2-5s, 5-10s, \textgreater 10s}, and present the average recall with 100 BMN proposals for each level.
As shown in Figure \ref{fig:error_time},
we make comparisons with ActivityNet, which shows  our dataset is more challenging in different durations due to its fine-grained and multi-label  annotations.
Moreover, instances in the \textit{0-2s} level have the worst performance,
even though this level covers the largest proportion of our FineAction dataset (Table \ref{tab:time}).
It clearly illustrates the localization difficulty in short instances.
Moreover, 
we further show the action categories with top-3 worst and top-3 best localization performance in Table  \ref{tab:error_class}.
We can see that,
the 3 worst categories (\textit{straighten hair}, \textit{apply eyebrows} and  \textit{dig hole}) are highly fine-grained with much shorter duration.
\textcolor{black}{
In Figure \ref{fig: apply_eyebrow}, we visualize the video clips for one challenging category of \textit{apply eyebrows}.
Note that, both on the spacial and temporal, apply eyebrows is a fine-grained action with short duration and slight movement.  }
Hence,
it is quite challenging to detect these actions,
even with a large number of training instances.

\begin{table}[t]
\centering
\begin{tabular}{c|ccc}
\hline
\textbf{FineAction}  & \textbf{AR@100} & \textbf{Avg. Duration} &\textbf{Num}\\
\hline
straighten hair        &   1.6    &  1.17 s &    1,934\\
apply eyebrows         &   1.7    &  1.13 s &    3,428\\
dig hole               &   2.2    &  0.83 s &     544\\
\hline
freestyle relay     &       88.4  &   53.12 s  &   345\\
breaststroke      &        94.5  &   53.59 s    &  211\\
play the harp       &     96.2   &  69.21 s   &   268\\
\hline
\end{tabular}
\caption{Top-3 worst and top-3 best localization performance of categories on the FineAction with BMN method. }
\label{tab:error_class}
\end{table}

\begin{table*}[t]
\centering 
\begin{minipage}{.48\textwidth}
\centering
\includegraphics[width=0.95\linewidth]{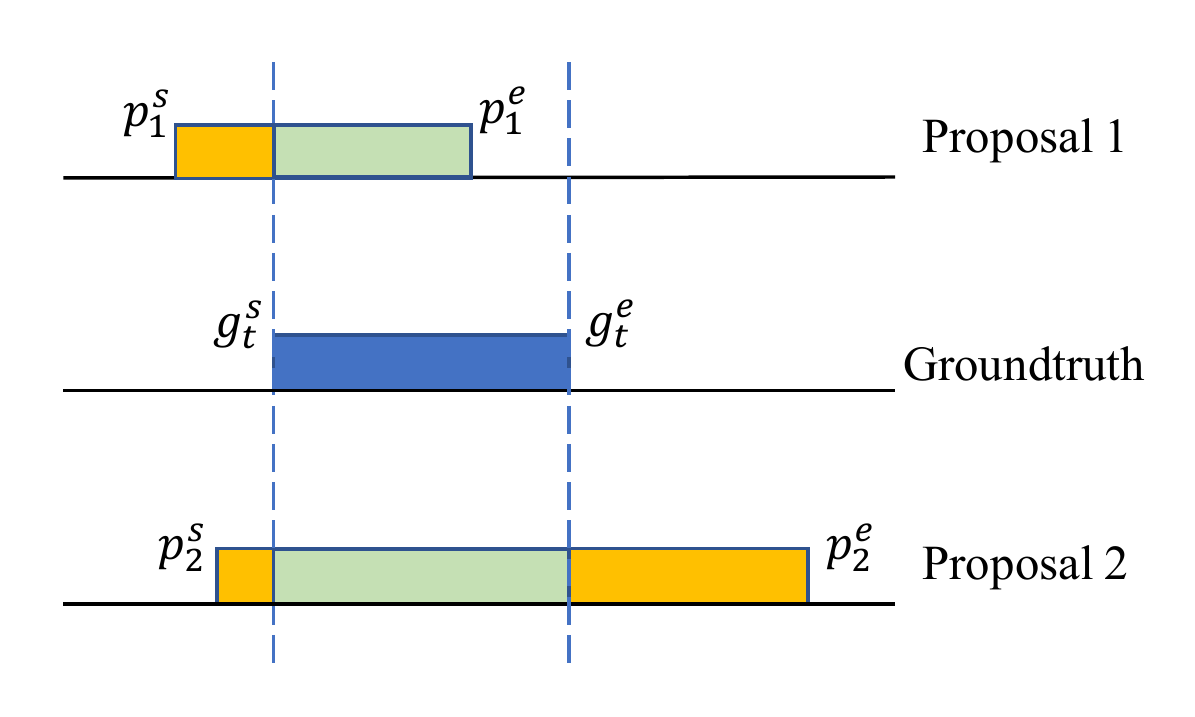}
\captionof{figure}{
An illustration of the proposed new metric. In this example, the IoU of the ground truth moment and two proposals (\textit{Proposal 1} and \textit{Proposal 2}) are both 0.5.}
\label{fig:metric}
\end{minipage}
\hfill
\begin{minipage}{.48\textwidth}
\centering
\begin{tabular}{c|ccc}
\hline
\textbf{Method}  & \textbf{Modality} & \textbf{AUC old} &\textbf{AUC new}\\
\hline
\multirow{3}{*}{BMN \cite{lin2019bmn}} & RGB & 17.49 & 16.42 \\
 & Flow & 18.94 & 17.94 \\
 & RGB+Flow & 19.19 & 18.19 \\
\hline
\multirow{3}{*}{DBG \cite{lin2020fast}} & RGB & 15.48 & 14.53\\
  & Flow & 17.70 & 16.74 \\
  & RGB+Flow & 17.24 & 16.32 \\
\hline
\multirow{3}{*}{G-TAD \cite{xu2020g}} & RGB & 16.06 & 14.90 \\
 & Flow & 17.09 & 15.97 \\
 & RGB+Flow & 17.65 & 16.54 \\
\hline
\end{tabular}
\caption{Comparison of different SOTA methods on new metric at tIoU [0.50:0.05:0.95]. Results are worse on the new metric, which can better evaluate the performance for shorter instance.}
\label{tab:new_metric}
\end{minipage}
\end{table*}

\textbf{Rethink the evaluation metrics for shorter instances.}
As mentioned in Section \ref{Evaluation}, we think that the prevailing evaluation metric AR@AN IoU is unreliable for shorter instances.
For the short ground truth,
it is preferable to generate a short proposal with the similar scale,
instead of producing a long proposal to cover the ground truth blindly.
For example, as shown in Figure \ref{fig:metric}, both \textit{Proposal 1} and \textit{Proposal 2} meet the IoU=0.5 condition, i.e., these two proposals are regarded equally under the original AR@AN IoU. 
However, since the temporal boundaries of the \textit{Proposal 2} are further from the ground truth, we propose a new metric as follows, which  penalizes more on \textit{Proposal 2}:
\begin{equation}
IoU_{new}=IoU\cdot \alpha_s \cdot \alpha_e,
\label{eq:new IoU}
\end{equation}
where $\alpha_s=1-\mathrm{abs}(g_t^s-p^s)$ and $\alpha_e=1-\mathrm{abs}(g_t^e-p^e)$.
Both $p^s,p^e$ and $g_t^s,g_t^e$ are normalized to 0-1 by dividing the whole video length.
As shown in Table \ref{tab:new_metric}, 
the new metric has lower performance than the old metric and is more suitable for fine-grained annotations. It also shows that action localization for shorter instances still faces with great challenges.

\section{Conclusion}
\label{conclusion}
In this paper, we have proposed a new large-scale and fine-grained video dataset, coined as FineAction, for temporal action localization.
FineAction differs from the existing datasets in aspects of fine-grained action definitions, high-quality and dense annotations, co-occurring instances of multiple classes.
We have empirically investigated several popular methods for temporal action localization on our FineAction and performed in-depth error analysis, which shows great challenges posed by our dataset for temporal action localization.
\textcolor{black}{
As for the future research, we think that fine-grained module design might improve the performance over FineAction.}
We hope our FineAction can facilitate new advances in the field of temporal action localization.


\bibliographystyle{IEEEtran}
\bibliography{Reference}

\end{document}